\documentclass[11pt]{article}
\usepackage{acl2014}
\usepackage{times}
\usepackage{url}
\usepackage{latexsym}
\usepackage{amsmath}
\usepackage{multirow}
\usepackage[normalem]{ulem}
\usepackage{color}
\usepackage[usenames,dvipsnames]{xcolor}
\usepackage{enumitem}
\usepackage[lined,boxed,commentsnumbered]{algorithm2e}
\usepackage{graphicx}
\usepackage{epstopdf}
\usepackage{subfig}
\usepackage{mdwlist}
\usepackage{enumitem}

\title{Improving Agreement and Disagreement Identification \\in Online Discussions with A Socially-Tuned Sentiment Lexicon}

\author{Lu Wang \\
  Department of Computer Science \\
  Cornell University \\
  Ithaca, NY 14853 \\
  {\tt luwang@cs.cornell.edu} \\\And
  Claire Cardie \\
  Department of Computer Science \\
  Cornell University \\
  Ithaca, NY 14853 \\
  {\tt cardie@cs.cornell.edu} \\}

\begin{document}
\maketitle

\begin{abstract}
We study the problem of agreement and disagreement detection in online discussions. An isotonic Conditional Random Fields (isotonic CRF) based sequential model is proposed to make predictions on sentence- or segment-level. We automatically construct a socially-tuned lexicon that is bootstrapped from existing general-purpose sentiment lexicons to further improve the performance. We evaluate our agreement and disagreement tagging model on two disparate online discussion corpora -- Wikipedia Talk pages and online debates. Our model is shown to outperform the state-of-the-art approaches in both datasets. For example, the isotonic CRF model achieves F1 scores of 0.74 and 0.67 for agreement and disagreement detection, when a linear chain CRF obtains 0.58 and 0.56 for the discussions on Wikipedia Talk pages.
\end{abstract}

\section{Introduction}
\label{sec:intro}
We are in an era where people can easily voice and exchange their opinions on the internet through forums or social media. Mining public opinion and the social interactions from online discussions is an important task, which has a wide range of applications. 
For example, by analyzing the users' attitude in forum posts on social and political problems, it is able to identify ideological stance~\cite{Somasundaran:2009:RSO:1687878.1687912} and user relations~\cite{qiu-yang-jiang:2013:NAACL-HLT}, and thus further discover subgroups~\cite{Hassan:2012:DSO:2390948.2390956,Abu-Jbara:2012:SDI:2390524.2390580} with similar ideological viewpoint. Meanwhile, catching the sentiment in the conversation can help detect online disputes, reveal popular or controversial topics, and potentially disclose the public opinion formation process.

\begin{figure}[ht]
\hspace{-3mm}
    {\small
    \begin{tabular}{|p{78mm}|}
    \hline
    \textbf{Zer0faults}: So questions comments feedback welcome. Other views etc. I just hope we can remove the assertations that WMD's were in fact the sole reason for the US invasion, considering that HJ Res 114 covers many many reasons.\\
    $>$\textbf{Mr. Tibbs}: So basically what you want to do is remove all mention of the cassus belli of the Iraq War and try to create the false impression that this military action was as inevitable as the sunrise.$_{[NN]}$ No. {\color{blue}Just because things didn't turn out the way the Bush administration wanted doesn't give you license to rewrite history.$_{[NN]}$} ...\\
    $>>$\textbf{MONGO}: Regardless, the article is an antiwar propaganda tool.$_{[NN]}$ ...\\
    $>>>$\textbf{Mr. Tibbs}: {\color{blue}So what?$_{[NN]}$} That wasn't the cassus belli and trying to give that impression After the Fact is Untrue.$_{[NN]}$ Hell, the reason it wasn't the cassus belli is because there are dictators in Africa that make Saddam look like a pussycat...\\
    $>>$\textbf{Haizum}: Start using the proper format or it's over for your comments.$_{[N]}$ {\color{blue}If you're going to troll, do us all a favor and stick to the guidelines.$_{[N]}$} ... \\
    \textbf{Tmorton166}: Hi, I wonder if, as an outsider to this debate I can put my word in here. I considered mediating this discussion however I'd prefer just to comment and leave it at that :). I agree mostly with what Zer0faults is saying$_{[PP]}$. ...\\
    $>$\textbf{Mr. Tibbs}: Here's the problem with that.$_{[NN]}$ It's not about publicity or press coverage. It's about the fact that the Iraq disarmament crisis set off the 2003 Invasion of Iraq. ... And theres a huge problem with rewriting the intro as if the Iraq disarmament crisis never happened.$_{[NN]}$\\
    $>>$\textbf{Tmorton166}: ... To suggest in the opening paragraph that the ONLY reason for the war was WMD's is wrong - because it simply isn't.$_{[NN]}$ However I agree that the emphasis needs to be on the armaments crisis because it was the reason sold to the public and the major one used to justify the invasion but it needs to acknowledge that there was at least 12 reasons for the war as well.$_{[PP]}$ ...\\
    \hline
	\end{tabular}
	}
	\vspace{-3mm}
    \caption{\small Example discussion from wikipedia talk page for article ``Iraq War", where editors discuss about the correctness of the information in the opening paragraph. We only show some sentences that are relevant for demonstration. Other sentences are omitted by ellipsis. Names of editors are in \textbf{bold}. ``$>$" is an indicator for the reply structure, where turns starting with $>$ are response for most previous turn that with one less $>$. We use ``\textit{NN}", ``\textit{N}", and ``\textit{PP}"  to indicate ``strongly disagree", ``disagree", and ``strongly agree". Sentences in {\color{blue}blue} are examples whose sentiment is hard to detect by an existing lexicon.}
    \label{fig:discussion_example}
\end{figure}

In this work, we study the problem of agreement and disagreement identification in online discussions. Sentence-level agreement and disagreement detection for this domain is challenging in its own right due to the dynamic nature of online conversations, and the less formal, and usually very emotional language used. As an example, consider a snippet of discussion from Wikipedia Talk page for article ``Iraq War" where editors argue on the correctness of the information in the opening paragraph (Figure~\ref{fig:discussion_example}). \textit{``So what?"} should presumably be tagged as a
negative sentence as should the sentence \textit{``If you're going to troll, do us all a favor and stick to the guidelines."}.
We hypothesize that these, and other, examples will be difficult for the tagger unless the context surrounding each sentence is considered and in the absence of a sentiment lexicon tuned for conversational text~\cite{Ding:2008:HLA,Choi:2009:APL}. 

As a result, we investigate isotonic Conditional Random Fields (isotonic CRF)~\cite{Mao+Lebanon:07a} for the sentiment tagging task since they preserve the advantages of the popular CRF sequential tagging models~\cite{Lafferty:2001:CRF} while providing an efficient mechanism to encode domain knowledge --- in our case, a sentiment lexicon --- through isotonic constraints on the model parameters. 
In particular, we bootstrap the construction of a sentiment lexicon from Wikipedia talk pages using the lexical items in existing general-purpose sentiment lexicons as seeds and in conjunction with an existing label propagation algorithm~\cite{citeulike:310457}.\footnote{Our online discussion lexicon (Section~\ref{sec:lexicon}) will be made publicly available.}

To summarize, our chief contributions include:

(1) We propose an agreement and disagreement identification model based on isotonic Conditional Random Fields~\cite{Mao+Lebanon:07a} to identify users' attitude in online discussion. Our predictions that are made on the sentence- or segment-level, are able to discover fine-grained sentiment flow within each turn, which can be further applied in other applications, such as dispute detection or argumentation structure analysis. 
We employ two existing online discussion data sets: the \textit{Authority and Alignment in Wikipedia Discussions
(AAWD)} corpus of \newcite{Bender:2011:ASA} (Wikipedia talk pages) and the \textit{Internet Argument Corpus (IAC)} of \newcite{WALKER12}. Experimental results show that our model significantly outperforms state-of-the-art methods on the AAWD data (our F1 scores are 0.74 and 0.67 for agreement and disagreement, vs. 0.58 and 0.56 for the linear chain CRF approach) and IAC data (our F1 scores are 0.61 and 0.78 for agreement and disagreement, vs. 0.28 and 0.73 for SVM).

(2) Furthermore, we construct a new sentiment lexicon for online discussion. We show that the learned lexicon significantly improves performance over systems that use existing general-purpose lexicons (i.e.\ MPQA lexicon~\cite{Wilson:2005:RCP}, General Inquirer~\cite{stone66}, and SentiWordNet~\cite{Esuli2006sentiwordnet}). Our lexicon is constructed from a very large-scale discussion corpus based on Wikipedia talk page, where previous work~\cite{Somasundaran:2010:RSI:1860631.1860645} for constructing online discussion lexicon relies on human annotations derived from limited number of conversations.

In the remainder of the paper, we describe first the related work (Section~\ref{sec:related}). Then we introduce the sentence-level agreement and disagreement identification model (Section~\ref{sec:model}) as well as the label propagation algorithm for lexicon construction (Section~\ref{sec:lexicon}). After explain the experimental setup, we display the results and provide further analysis in Section~\ref{sec:result}.

\section{Related Work}
\label{sec:related}
Sentiment analysis has been utilized as a key enabling technique in a number of conversation-based applications. Previous work mainly studies the attitudes in spoken meetings~\cite{GalleyEtal,hahn-ladner-ostendorf:2006:HLT-NAACL06-Short} or broadcast conversations~\cite{Wang:2011:DAD} using Conditional Random Fields (CRF)~\cite{Lafferty:2001:CRF}. \newcite{GalleyEtal} employ Conditional Markov models to detect if discussants reach at an agreement in spoken meetings. Each state in their model is an individual turn and prediction is made on the turn-level. In the same spirit, \newcite{Wang:2011:DAD} also propose a sequential model based on CRF for detecting agreements and disagreements in broadcast conversations, where they primarily show the efficiency of prosodic features. While we also exploit a sequential model extended from CRFs, our predictions are made for each sentence or segment rather than at the turn-level. Moreover, we experiment with online discussion datasets that exhibit a more realistic distribution of disagreement vs.\ agreement, where much more disagreement is observed due to its function and the relation between the participants. This renders the detection problem more challenging.

Only recently, agreement and disagreement detection is studied for online discussion, especially for online debate. \newcite{Abbott:2011:YST} investigate different types of features based on dependency relations as well as {\it manually}-labeled features, such as if the participants are nice, nasty, or sarcastic, and respect or insult the target participants. Automatically inducing those features from human annotation are challenging itself, so it would be difficult to reproduce their work on new datasets.
We use only automatically generated features. Using the same dataset, \newcite{misra-walker:2013:SIGDIAL} study the effectiveness of topic-independent features, e.g. discourse cues indicating agreement or negative opinion. Those cues, which serve a similar purpose as a sentiment lexicon, are also constructed manually. In our work, we create an online discussion lexicon automatically and construct sentiment features based on the lexicon. Also targeting online debate, \newcite{Yin:2012:ULG:2392963.2392978} train a logistic regression classifier with features aggregating posts from the same participant to predict the sentiment for each individual post. This approach works only when the speaker has enough posts on each topic, which is not applicable to newcomers. \newcite{Hassan:2010:WAI:1870658.1870779} focus on predicting the attitude of participants towards each other. They relate the sentiment words to the second person pronoun, which produces strong baselines. We also adopt their baselines in our work. Although there are available datasets with (dis)agreement annotated on Wikipedia talk pages, we are not aware of any published work that utilizes these annotations. Dialogue act recognition on talk pages~\cite{Ferschke2012BAR} might be the most related.

While detecting agreement and disagreement in conversations is useful on its own, it is also a key component for related tasks, such as stance prediction~\cite{Thomas:2006:GOV:1610075.1610122,Somasundaran:2009:RSO:1687878.1687912,WalkerAAG12} and subgroup detection~\cite{Hassan:2012:DSO:2390948.2390956,Abu-Jbara:2012:SDI:2390524.2390580}.
For instance, \newcite{Thomas:2006:GOV:1610075.1610122} train an agreement detection classifier with Support Vector Machines on congressional floor-debate transcripts to determine whether the speeches represent support of or opposition to the proposed legislation. \newcite{Somasundaran:2009:RSO:1687878.1687912} design various sentiment constraints for inclusion in an integer linear programming framework for stance classification. For subgroup detection, \newcite{Abu-Jbara:2012:SDI:2390524.2390580} uses the polarity of the expressions in the discussions and partition discussants into subgroups based on the intuition that people in the same group should mostly agree with each other. Though those work highly relies on the component of agreement and disagreement detection, the evaluation is always performed on the ultimate application only.

\section{The Model}
\label{sec:model}
We first give a brief overview on isotonic Conditional Random Fields (isotonic CRF)~\cite{Mao+Lebanon:07a}, which is used as the backbone approach for our sentence- or segment-level agreement and disagreement detection model. 
We defer the explanation of online discussion lexicon construction in Section~\ref{sec:lexicon}.

\subsection{Problem Description}
Consider a discussion comprised of sequential turns uttered by the participants; each turn consists of a sequence of text units, where each unit can be a sentence or a segment of several sentences. Our model takes as input the text units $\mathbf{x}=\{x_{1}, \cdots, x_{n}\}$ in the same turn, and outputs a sequence of sentiment labels $\mathbf{y}=\{y_{1}, \cdots, y_{n}\}$, where $y_{i}\in \mathcal{O},  \mathcal{O}=\{\mathrm{NN, N, O, P, PP}\}$. The labels in $\mathcal{O}$ represent strongly disagree (NN), disagree (N), neutral (O), agree (P), strongly agree (PP), respectively. In addition, elements in the partially ordered set $\mathcal{O}$ possess an ordinal relation $\leq$. Here, we differentiate agreement and disagreement with different intensity, because the output of our classifier can be used for other applications, such as dispute detection, where ``strongly disagree" (e.g. NN) plays an important role. Meanwhile, fine-grained sentiment labels potentially provide richer context information for the sequential model employed for this task.

\subsection{Isotonic Conditional Random Fields}
\label{subsec:isotonic}
Conditional Random Fields (CRF) have been successfully applied in numerous sequential labeling tasks~\cite{Lafferty:2001:CRF}. Given a sequence of utterances or segments $\mathbf{x}=\{x_{1}, \cdots, x_{n}\}$, according to linear-chain CRF, the probability of the labels $\mathbf{y}$ for $\mathbf{x}$ is given by:

{
\begin{equation}
\begin{aligned}
p(\mathbf{y}|\mathbf{x}) = & \frac{1}{Z(\mathbf{x})}\exp (\sum_{i}\sum_{\sigma, \tau} \lambda_{\langle \sigma, \tau \rangle} f_{\langle \sigma, \tau \rangle}(y_{i-1}, y_{i})\\
& + \sum_{i}\sum_{\sigma, w} \mu_{\langle \sigma, w \rangle} g_{\langle \sigma, w \rangle}(y_{i}, x_{i}))\\
\end{aligned}
\end{equation}
}

$f_{\langle \sigma, \tau \rangle}(y_{i-1}, y_{i})$ and $g_{\langle \sigma, w \rangle}(y_{i}, x_{i})$ are feature functions. Given that $y_{i-1}, y_{i}, x_{i}$ take values of $\sigma, \tau, w$, the functions are indexed by pairs $\langle \sigma, \tau \rangle$ and $\langle \sigma, w \rangle$. $\lambda_{\langle \sigma, \tau \rangle}, \mu_{\langle \sigma, w \rangle}$ are the parameters.

CRF, as defined above, is not appropriate for ordinal data like sentiment, because it ignores the ordinal relation among sentiment labels. Isotonic Conditional Random Fields (isotonic CRF) are proposed by \newcite{Mao+Lebanon:07a} to enforce a set of monotonicity constraints on the parameters that are consistent with the ordinal structure and domain knowledge (in our case, a sentiment lexicon automatically constructed from online discussions).

Given a lexicon $\mathcal{M}=\mathcal{M}_{p}\cup \mathcal{M}_{n}$, where $\mathcal{M}_{p}$ and $\mathcal{M}_{n}$ are two sets of features (usually words) identified as strongly associated with positive sentiment and negative sentiment. The constraints are encoded as below. For each feature $w \in \mathcal{M}_{p}$, isotonic CRF enforces $\sigma \leq \sigma^{\prime} \Rightarrow \mu_{\langle \sigma, w \rangle} \leq \mu_{\langle \sigma^{\prime}, w \rangle}$. Intuitively, the parameters $\mu_{ \langle \sigma, w \rangle }$ are intimately tied to the model probabilities. When a feature such as ``totally agree" is observed in the training data, the feature parameter for $\mu_{\langle \mathrm{PP}, \mathrm{totally~agree} \rangle }$ is likely to increase. Similar constraints are also defined on $\mathcal{M}_{n}$. In this work, we boostrap the construction of an online discussion sentiment lexicon used as $\mathcal{M}$ in the isotonic CRF (see Section~\ref{sec:lexicon}).

The parameters can be found by maximizing the likelihood subject to the monotonicity constraints. We adopt the re-parameterization from~\newcite{Mao+Lebanon:07a} for a simpler optimization problem, and refer the readers to~\newcite{Mao+Lebanon:07a} for more details.\footnote{The full implementation is based on MALLET~\cite{McCallumMALLET}. We thank Yi Mao for sharing the implementation of the core learning algorithm.}

\subsection{Features}
\label{sub:sentimentfeatures}

\begin{table}
    {\fontsize{10}{12}\selectfont
    \begin{tabular}{|l|}
    \hline
    
    \underline{\bf Lexical Features} \\
    - unigram/bigram \\
    - num of words all uppercased \\
    - num of words \\
    
    \underline{{\bf Discourse Features}} \\
	- initial uni-/bi-/trigram \\
	- repeated punctuations \\	
	- hedging~\cite{Farkas:2010:CST:1870535.1870536} \\
	- number of negators \\
    
	\underline{{\bf Syntactic/Semantic Features}} \\     
    - unigram with POS tag \\
	- dependency relation \\
	
	\underline{{\bf Conversation Features}} \\ 
	- quote overlap with target \\ 
    - TFIDF similarity with target (remove quote first) \\
    
    \underline{{\bf Sentiment Features}} \\
    - connective + sentiment words \\
    - sentiment dependency relation \\
    - sentiment words \\

	\hline
    \end{tabular}    
    }
    \caption{Features used in sentiment prediction.}
    \label{tab:feature_isocrf}
\end{table}

The features used in sentiment prediction are listed in Table~\ref{tab:feature_isocrf}. 
Features with numerical values are first normalized by standardization, then binned into 5 categories.

\paragraph{Syntactic/Semantic Features.}
Dependency relations have been shown to be effective for various sentiment prediction tasks~\cite{Joshi:2009:GDF:1667583.1667680,Somasundaran:2009:RSO:1687878.1687912,Hassan:2010:WAI:1870658.1870779,Abu-Jbara:2012:SDI:2390524.2390580}. 
We have two versions of dependency relation as features, one being the original form, another generalizing a word to its POS tag in turn. For instance, ``nsubj(wrong, you)" is generlized as the ``nsubj(\texttt{ADJ}, you)" and ``nsubj(wrong, \texttt{PRP})". We use Stanford parser~\cite{stanford_dependencies} to obtain parse trees and dependency relations.

\paragraph{Discourse Features.}
Previous work~\cite{Hirschberg:1993:ESD:972487.972490,Abbott:2011:YST} suggests that discourse markers, such as \textit{what?}, \textit{actually}, may have their use for expressing opinions. 
We extract the initial unigram, bigram, and trigram of each utterance as discourse features~\cite{Hirschberg:1993:ESD:972487.972490}. 
Hedge words are collected from the CoNLL-2012 shared task~\cite{Farkas:2010:CST:1870535.1870536}.

\paragraph{Conversation Features.}
Conversation features encode some useful information regarding the similarity between the current utterance(s) and the sentences uttered by the target participant. TFIDF similarity is computed. We also check if the current utterance(s) quotes target sentences and compute its length.

\paragraph{Sentiment Features.}
We gather connectives from Penn Discourse TreeBank~\cite{penntreediscourse} and combine them with any sentiment word that precedes or follows it as new features. Sentiment dependency relations are the subset of dependency relations with sentiment words. We replace those words with their polarity equivalents. For example, relation ``nsubj(wrong, you)" becomes ``nsubj(\texttt{SentiWord}$_{neg}$, you)".

\section{Online Discussion Sentiment Lexicon Construction}
\label{sec:lexicon}
\begin{table*}[ht]
\centering
    {
    \begin{tabular}{|p{160mm}|}
    \hline
	\textsc{Positive}\\ \hline
	please elaborate, nod, await response, from experiences, anti-war, profits, promises of, is undisputed, royalty, sunlight, conclusively, badges, prophecies, in vivo, tesla, pioneer, published material, from god, plea for, lend itself, geek, intuition, morning, anti \texttt{SentiWord}$_{neg}$, connected closely, Rel(undertake, to), intelligibility, Rel(articles, detailed), of noting, for brevity, Rel(believer, am), endorsements, testable, source carefully\\ \hline
	\textsc{Negative} \\ \hline
	: (, TOT, ?!!, in contrast, ought to, whatever, Rel(nothing, you), anyway, Rel(crap, your), by facts, purporting, disproven, Rel(judgement, our), Rel(demonstrating, you), opt for, subdue to, disinformation, tornado, heroin, Rel(newbies, the), Rel (intentional, is), pretext, watergate, folly, perjury, Rel(lock, article), contrast with, poke to, censoring information, partisanship, insurrection, bigot, Rel(informative, less), clowns, Rel(feeling, mixed), never-ending\\ \hline
	\end{tabular}
	}
    \caption{Example terms and relations from our online discussion lexicon. We choose for display terms that do not contain any seed word.}
    \label{tab:lexicon_example}
\end{table*}

So far as we know, there is no lexicon available for online discussions. Thus, we create from a large-scale corpus via \textit{label propagation}.
The label propagation algorithm, proposed by~\newcite{citeulike:310457}, is a semi-supervised learning method. In general, it takes as input a set of seed samples (e.g. sentiment words in our case), and the similarity between pairwise samples, then iteratively assigns values to the unlabeled samples (see Algorithm~\ref{alg:label_prop}).
The construction of graph $G$ is discussed in Section~\ref{subsec:graph}. 
Sample sentiment words in the new lexicon are listed in Table~\ref{tab:lexicon_example}.

\begin{algorithm}
\SetKwInOut{Input}{Input}
\SetKwInOut{Output}{Output}
\Input{$G=(V,E), w_{ij}\in [0, 1]$, positive seed words $P$, negative seed words $N$, number of iterations $T$}
\Output{\{$y_{i}$\}$_{i=0}^{\mid V \mid-1}$}
\BlankLine
$y_{i}=1.0$, $\forall v_{i}\in P$\\
$y_{i}=-1.0$, $\forall v_{i}\in N$\\
$y_{i}=0.0$, $\forall v_{i} \notin P\cup N$\\
\BlankLine

\For{$t=1 \cdots T$}{
	$y_{i}=\frac{\sum_{(v_{i}, v_{j})\in E} w_{ij} \times y_{j}}{\sum_{(v_{i}, v_{j})\in E} w_{ij}}$, $\forall v_{i}\in V$\\
	$y_{i}=1.0$, $\forall v_{i}\in P$\\
	$y_{i}=-1.0$, $\forall v_{i}\in N$\\
}

\caption{The label propagation algorithm~\cite{citeulike:310457} used for constructing online discussion lexicon.}
\label{alg:label_prop}
\end{algorithm}

\subsection{Graph Construction}
\label{subsec:graph}

\paragraph{Node Set $V$.}
Traditional lexicons, like General Inquirer~\cite{stone66}, usually consist of polarized unigrams. 
As we mentioned in Section~\ref{sec:intro}, unigrams lack the capability of capturing the sentiment conveyed in online discussions. Instead, bigrams, dependency relations, and even punctuation can serve as supplement to the unigrams. Therefore, we consider four types of \textit{text units} as nodes in the graph: unigrams, bigrams, dependency relations, sentiment dependency relations. Sentiment dependency relations are described in Section~\ref{sub:sentimentfeatures}. We replace all relation names with a general label.
Text units that appear in at least 10 discussions are retained as nodes to reduce noise.

\paragraph{Edge Set $E$.}
As \newcite{Velikovich2010VWP} and \newcite{conf/acl/FengKKC13} notice, a dense graph with a large number of nodes is susceptible to propagating noise, and will not scale well. We thus adopt the algorithm in \newcite{conf/acl/FengKKC13} to construct a sparsely connected graph. For each text unit $t$, we first compute its representation vector $\vec a$ using Pairwise Mutual Information scores with respect to the top 50 co-occuring text units. We define ``co-occur" as text units appearing in the same sentence. An edge is created between two text units $t_{0}$ and $t_{1}$ only if they ever co-occur. The similarity between $t_{0}$ and $t_{1}$ is calculated as the Cosine similarity between $\vec a_{0}$ and $\vec a_{1}$.

\paragraph{Seed Words.}
The seed sentiment are collected from three existing lexicons: MPQA lexicon, General Inquirer, and SentiWordNet. Each word in SentiWordNet is associated with a positive score and a negative score; words with a polarity score larger than 0.7 are retained. We remove words with conflicting sentiments.


\subsection{Data}
The graph is constructed based on Wikipedia talk pages. We download the 2013-03-04 Wikipedia data dump, which contains 4,412,582 talk pages. Since we are interested in conversational languages, we filter out talk pages with fewer than 5 participants. This results in a dataset of 20,884 talk pages, from which the graph is constructed.

\section{Experimental Setup}
\label{sec:expsetup}
\subsection{Datasets}
\paragraph{Wikipedia Talk pages.}
The first dataset we use is \textit{Authority and Alignment in Wikipedia Discussions (AAWD)} corpus~\cite{Bender:2011:ASA}.
AAWD consists of 221 English Wikipedia discussions with agreement and disagreement annotations.\footnote{\newcite{Bender:2011:ASA} originally use positive alignment and negative alignment to indicate two types of social moves. They define those alignment moves as ``agreeing or disagreeing" with the target. We thus use agreement and disagreement instead of positive and negative alignment in this work.} 

The annotation of AAWD is made at utterance- or turn-level, where a turn is defined as continuous body of text uttered by the same participant. Annotators either label each utterance as agreement, disagreement or neutral, and select the corresponding spans of text, or label the full turn. Each turn is annotated by two or three people. To induce an utterance-level label for instances that have only a turn-level label, we assume they have the same label as the turn.

To train our sentiment model, we further transform agreement and disagreement labels (i.e. 3-way) into the 5-way labels. For utterances that are annotated as agreement and have the text span specified by at least two annotators, they are treated as ``strongly agree" (PP). If an utterance is only selected as agreement by one annotator or it gets the label by turn-level annotation, it is ``agree" (P). ``Strongly disagree" (NN) and ``disagree" (N) are collected in the same way from disagreement label. All others are neutral (O). 
In total, we have 16,501 utterances. 1,930 and 1,102 utterances are labeled as ``NN" and ``N". 532 and 99 of them are ``PP" and ``P". All other 12,648 are neutral samples.~\footnote{345 samples with both positive and negative labels are treated as neutral.}

\paragraph{Online Debate.}
The second dataset is the \textit{Internet Argument Corpus (IAC)}~\cite{WALKER12} collected from an online debate forum. Each discussion in IAC consists of multiple posts, where we treat each post as a turn. Most posts (72.3\%) contain quoted content from the posts they target at or other resources. A post can have more than one quote, which naturally break the post into multiple segments. 1,806 discussions are annotated with agreement and disagreement on the segment-level from -5 to 5, with -5 as strongly disagree and 5 as strongly agree.
We first compute the average score for each segment among different annotators and transform the score into sentiment label in the following way. We treat $[-5, -3]$ as NN (1595 segments), $(-3, -1]$ as N (4548 segments), $[1, 3)$ as P (911 samples), $[3, 5]$ as PP (199), all others as O (290 segments). 

In the test phase, utterances or segments predicted with NN or N are treated as disagreement; the ones predicted as PP or P are agreement; O is neutral.


\subsection{Comparison}
We compare with two baselines. (1) \textbf{Baseline (Polarity)} is based on counting the sentiment words from our lexicon. An utterance or segment is predicted as agreement if it contains more positive words than negative words, or disagreement if more negative words are observed. Otherwise, it is neutral. (2) \textbf{Baseline (Distance)} is extended from~\cite{Hassan:2010:WAI:1870658.1870779}. Each sentiment word is associated with the closest second person pronoun, and a surface distance can be computed between them. A classifier based on Support Vector Machines~\cite{Joachims:1999:MLS:299094.299104} (SVM) is trained with the features of sentiment words, minimum/maximum/average of the distances.

We also compare with two state-of-the-art methods that are widely used in sentiment prediction for conversations. The first one is an RBF kernel SVM based approach,  which has been used for sentiment prediction~\cite{Hassan:2010:WAI:1870658.1870779}, and (dis)agreement detection~\cite{Yin:2012:ULG:2392963.2392978} in online debates.
The second is linear chain CRF, which has been utilized for (dis)agreement identification in broadcast conversations~\cite{Wang:2011:DAD}.

\section{Results}
\label{sec:result}
\begin{table*}[ht]
\centering
    {\fontsize{10}{11}\selectfont
	\begin{tabular}{|l|ccc|ccc|}
    \hline
	&\multicolumn{3}{|c|}{\textbf{Strict F1}}&\multicolumn{3}{|c|}{\textbf{Soft F1}}\\\hline
	& \textbf{Agree} & \textbf{Disagree} & \textbf{Neutral} & \textbf{Agree} & \textbf{Disagree} & \textbf{Neutral} \\ \hline
	Baseline (Polarity) & 14.56 & 25.70 & 64.04 & 22.53 & 38.61 & 66.45\\ 
	Baseline (Distance) & 8.08 & 20.68 & 84.87 & 33.75 & 55.79 & 88.97 \\ \hline
	SVM (3-way) & 26.76& 35.79& 77.39& 44.62 & 52.56 & 80.84\\ 
	~~~~ + downsampling & ~~~~ 21.60 & ~~~~ 36.32 & ~~~~ 72.11 & ~~~~ 31.86 & ~~~~ 49.58 & ~~~~ 74.92\\ 
	CRF (3-way) & 20.99& 23.85& 85.28 & 56.28 & 56.37& 89.41\\ 
	CRF (5-way) & 20.47& 19.42& 85.86 & 58.39& 56.30 & 90.10\\ 
	~~~~ + downsampling & ~~~~ 24.26& ~~~~ 31.28& ~~~~ 77.12& ~~~~ 47.30& ~~~~ 46.24& ~~~~ 80.18\\ \hline
	isotonic CRF & 24.32& 21.95& 86.26& 68.18& 62.53& 88.87\\ 
	~~~~ + downsampling & ~~~~ 29.62& ~~~~ 34.17& ~~~~ 80.97 & ~~~~ 55.38 & ~~~~ 53.00 & ~~~~ 84.56\\ 
	~~~~ + new lexicon & \textbf{46.01} & \textbf{\textit{51.49}} & \textit{87.40} & \textbf{\textit{74.47}} & \textbf{\textit{67.02}} & \textit{90.56}\\ 
	~~~~ + new lexicon + downsampling& ~~~~ \textbf{\textit{47.90}} & ~~~~ \textbf{49.61} & ~~~~ 81.60& ~~~~ 64.97& ~~~~ 58.97 & ~~~~ 84.04\\ \hline

	\end{tabular}
	}
    \caption{Strict and soft F1 scores for agreement and disagreement detection on Wikipedia talk pages (AAWD). All the numbers are multiplied by 100. In each column, \textbf{bold} entries (if any) are statistically significantly higher than all the rest, and the \textit{italic} entry has the highest absolute value.
    Our model based on the isotonic CRF with the new lexicon produces significantly better results than all the other systems for agreement and disagreement detection. Downsampling, however, is not always helpful.}
	\label{tab:aawd_main}
\end{table*}

In this section, we first show the experimental results on sentence- and segment-level agreement and disagreement detection in two types of online discussions -- \textit{Wikipedia Talk pages} and \textit{online debates}. Then we provide more detailed analysis for the features used in our model. Furthermore, we discuss several types of errors made in the model.

\subsection{Wikipedia Talk Pages}

We evaluate the systems by standard F1 score on each of the three categories: agreement, disagreement, and neutral. For AAWD, we compute two versions of F1 scores. \textbf{Strict F1} is computed against the true labels. For \textbf{soft F1}, if a sentence is never labeled by any annotator on the sentence-level and adopts its agreement/disagreement label from the turn-level annotation, then it is treated as a true positive when predicted as neutral.

\begin{table}[ht]
    {\fontsize{8}{9}\selectfont
	\begin{tabular}{|l|ccc|}
    \hline
	&\textbf{Agree} & \textbf{Disagree} & \textbf{Neu}\\ \hline
	Baseline (Polarity) & 3.33& 5.96 & 65.61\\ 
	Baseline (Distance) & 1.65 & 5.07 & 85.41\\ \hline
	SVM (3-way) & 25.62& 69.10 & 31.47\\ 
	~~~~ + new lexicon features & ~~~~ 28.35& ~~~~ 72.58& ~~~~ 34.53\\ 
	CRF (3-way) & 29.46& 74.81& 31.93\\ 
	CRF (5-way) & 24.54& 69.31& 39.60\\ 
	~~~~ + new lexicon features & ~~~~ 28.85& ~~~~ 71.81& ~~~~ 39.14\\ \hline
	isotonic CRF & \textbf{53.40} & \textbf{76.77}& \textit{44.10}\\ 
	~~~~ + new lexicon  & ~~~~ \textbf{\textit{61.49}} & ~~~~ \textbf{\textit{77.80}}& ~~~~ \textit{51.43}\\ \hline

	\end{tabular}
	}
    \caption{F1 scores for agreement and disagreement detection on online debate (IAC). All the numbers are multiplied by 100. In each column, \textbf{bold} entries (if any) are statistically significantly higher than all the rest, and the \textit{italic} entry has the highest absolute value except baselines.
    We have two main observations: 1) Both of our models based on isotonic CRF significantly outperform other systems for agreement and disagreement detection. 2) By adding the new lexicon, either as features or constraints in isotonic CRF, all systems achieve better F1 scores.}
    \label{tab:iac_main}

\end{table}

Table~\ref{tab:aawd_main} demonstrates our main results on the Wikipedia Talk pages (AAWD dataset). Without downsampling, our isotonic CRF based systems with the new lexicon significantly outperform the compared approaches for agreement and disagreement detection according to the paired-$t$ test ($p<0.05$). We also perform downsampling by removing the turns only containing neutral utterances. However, it does not always help with performance. We suspect that, with less neutral samples in the training data, the classifier is less likely to make neutral predictions, which thus decreases true positive predictions. For strict F-scores on agreement/disagreement, downsampling has mixed effect, but mostly we get slightly better performance.

\subsection{Online Debates}

Similarly, F1 scores for agreement, disagreement and neutral for online debates (IAC dataset) are displayed in Table~\ref{tab:iac_main}. Both of our systems based on isotonic CRF achieve significantly better F1 scores than the comparison. Especially, our system with the new lexicon produces the best results. For SVM and linear-chain CRF based systems, we also add new sentiment features constructed from the new lexicon as described in Section~\ref{sub:sentimentfeatures}. We can see that those sentiment features also boost the performance for both of the compared approaches. 

\subsection{Feature Evaluation}
Moreover, we evaluate the effectiveness of features by adding one type of features each time. The results are listed in Table~\ref{tab:aawd_diff_featureset}. As it can be seen, the performance gets improved incrementally with every new set of features.

We also utilize $\chi^{2}$-test to highlight some of the salient features on the two datasets. We can see from Table~\ref{tab:feature_analysis} that, for online debates (IAC), some features are highly topic related, such as ``\textit{the male}" or ``\textit{the scientist}". This observation concurs with the conclusion in~\newcite{misra-walker:2013:SIGDIAL} that features with topic information are indicative for agreement and disagreement detection.

\begin{table}[ht]

    {\fontsize{8}{9}\selectfont
	\begin{tabular}{|l|ccc|}
    \hline

	\textbf{AAWD} & \textbf{Agree} & \textbf{Disagree} & \textbf{Neu}\\ \hline
	Lex & 40.77 & 52.90 & 79.65\\ \hline
	Lex + Syn & 68.18 & 63.91 & 88.87 \\ \hline
	Lex + Syn + Disc  & 70.93 & 63.69& 89.32\\ \hline
	Lex + Syn + Disc + Con  &  71.27& 63.72 &  89.60\\ \hline
	Lex + Syn + Disc + Con + Sent  & \textbf{74.47} & \textbf{67.02} & 90.56\\ \hline
	\end{tabular}
	}
	
	\vspace{5mm}
	{\fontsize{8}{9}\selectfont
	\begin{tabular}{|l|ccc|}
    \hline
	\textbf{IAC} &\textbf{Agree} & \textbf{Disagree} & \textbf{Neu}\\ \hline
	Lex & 56.65& 75.35& 45.72\\ \hline
	Lex + Syn & 54.16& 75.13& 46.12\\ \hline
	Lex + Syn + Disc & 54.27& 76.41& 47.60\\ \hline
	Lex + Syn + Disc + Con & 55.31& 77.25& 48.87\\ \hline
	Lex + Syn + Disc + Con + Sent  & \textbf{61.49} & 77.80 & \textbf{51.43}\\ \hline
	\end{tabular}
	}
    \caption{Results on Wikipedia talk page (AAWD) (with soft F1 score) and online debate (IAC) with different feature sets (i.e \textbf{Lex}ical, \textbf{Syn}tacitc/Semantic, \textbf{Disc}ourse, \textbf{Con}versation, and \textbf{Sent}iment features) by using isotonic CRF. The numbers in \textbf{bold} are statistically significantly higher than the numbers above it (paired-$t$ test, $p<0.05$).}
    \label{tab:aawd_diff_featureset}

\end{table}

%

\begin{table}[ht]

    {\fontsize{10}{11}\selectfont
    \begin{tabular}{|p{72mm}|}
    \hline
    \textbf{AAWD} \\
	\underline{\textsc{Positive}}:
	agree, nsubj (agree, I), nsubj (right, you), Rel (Sentiment$_{pos}$, I), thanks, amod (idea, good), nsubj(glad, I), good point, concur, happy with, advmod (good, pretty), suggestion$_{Hedge}$\\ 
	
	\underline{\textsc{Negative}}: 
	you, your, nsubj (negative, you), numberOfNegator, don't, nsubj (disagree, I), actually$_{SentInitial}$, please stop$_{SentInitial}$, what ?$_{SentInitial}$, should$_{Hedge}$\\ \hline
	\end{tabular}
	}
	
	{\fontsize{10}{11}\selectfont
	\begin{tabular}{|p{72mm}|}
    \hline
    \textbf{IAC} \\
	\underline{\textsc{Positive}}:
	amod (conclusion, logical), Rel (agree, on), Rel (have, justified), Rel (work, out), one might$_{SentInitial}$, to confirm$_{Hedge}$, women\\
	
	\underline{\textsc{Negative}}:
	their kind, the male, the female, the scientist, according to, is stated, poss (understanding, my), hell$_{SentInitial}$, whatever$_{SentInitial}$\\ \hline
	\end{tabular}
	}
    \caption{Relevant features by $\chi^{2}$ test on AAWD and IAC.}
    \label{tab:feature_analysis}

\end{table}

%

\subsection{Error Analysis}
After a closer look at the data, we found two major types of errors. Firstly, people express disagreement not only by using opinionated words, but also by providing contradictory example. This needs a deeper understanding of the semantic information embedded in the text. Techniques like textual entailment can be used in the further work. 
Secondly, a sequence of sentences with sarcasm is hard to detect. For instance, \textit{``Bravo, my friends! Bravo! Goebbles would be proud of your abilities to whitewash information."} We observe terms like ``Bravo", ``friends", and ``be proud of" that are indicators for positive sentiment; however, they are in sarcastic tone. We believe a model that is able to detect sarcasm would further improve the performance.

\section{Conclusion}
We present an agreement and disagreement detection model based on isotonic CRFs that outputs labels at the sentence- or segment-level. We bootstrap the construction of a sentiment lexicon for online discussions, encoding it in the form of domain knowledge for the isotonic CRF learner. Our sentiment-tagging model is shown to outperform the state-of-the-art approaches on both Wikipedia Talk pages and online debates.

\vspace*{5mm}
\noindent{}
{\bf Acknowledgments} We heartily thank the Cornell NLP Group and the reviewers for helpful comments. This work was supported in part by NSF grants IIS-0968450 and IIS-1314778, and DARPA DEFT Grant FA8750-13-2-0015. The views and conclusions contained herein are those of the authors and should not be interpreted as necessarily representing the official policies or endorsements, either expressed or implied, of NSF, DARPA or the U.S. Government.

\bibliographystyle{acl}

\begin{thebibliography}{}

\bibitem[\protect\citename{Abbott \bgroup et al.\egroup }2011]{Abbott:2011:YST}
Rob Abbott, Marilyn Walker, Pranav Anand, Jean~E. Fox~Tree, Robeson Bowmani,
  and Joseph King.
\newblock 2011.
\newblock How can you say such things?!?: Recognizing disagreement in informal
  political argument.
\newblock In {\em Proceedings of the Workshop on Languages in Social Media},
  LSM '11, pages 2--11, Stroudsburg, PA, USA. Association for Computational
  Linguistics.

\bibitem[\protect\citename{Abu-Jbara \bgroup et al.\egroup
  }2012]{Abu-Jbara:2012:SDI:2390524.2390580}
Amjad Abu-Jbara, Mona Diab, Pradeep Dasigi, and Dragomir Radev.
\newblock 2012.
\newblock Subgroup detection in ideological discussions.
\newblock In {\em Proceedings of the 50th Annual Meeting of the Association for
  Computational Linguistics: Long Papers - Volume 1}, ACL '12, pages 399--409,
  Stroudsburg, PA, USA. Association for Computational Linguistics.

\bibitem[\protect\citename{Bender \bgroup et al.\egroup }2011]{Bender:2011:ASA}
Emily~M. Bender, Jonathan~T. Morgan, Meghan Oxley, Mark Zachry, Brian
  Hutchinson, Alex Marin, Bin Zhang, and Mari Ostendorf.
\newblock 2011.
\newblock Annotating social acts: Authority claims and alignment moves in
  wikipedia talk pages.
\newblock In {\em Proceedings of the Workshop on Languages in Social Media},
  LSM '11, pages 48--57, Stroudsburg, PA, USA. Association for Computational
  Linguistics.

\bibitem[\protect\citename{Choi and Cardie}2009]{Choi:2009:APL}
Yejin Choi and Claire Cardie.
\newblock 2009.
\newblock Adapting a polarity lexicon using integer linear programming for
  domain-specific sentiment classification.
\newblock In {\em Proceedings of the 2009 Conference on Empirical Methods in
  Natural Language Processing: Volume 2 - Volume 2}, EMNLP '09, pages 590--598,
  Stroudsburg, PA, USA. Association for Computational Linguistics.

\bibitem[\protect\citename{de Marneffe \bgroup et al.\egroup
  }2006]{stanford_dependencies}
Marie-Catherine de~Marneffe, Bill MacCartney, and Christopher~D. Manning.
\newblock 2006.
\newblock Generating typed dependency parses from phrase structure trees.
\newblock In {\em LREC}.

\bibitem[\protect\citename{Ding \bgroup et al.\egroup }2008]{Ding:2008:HLA}
Xiaowen Ding, Bing Liu, and Philip~S. Yu.
\newblock 2008.
\newblock A holistic lexicon-based approach to opinion mining.
\newblock In {\em Proceedings of the 2008 International Conference on Web
  Search and Data Mining}, WSDM '08, pages 231--240, New York, NY, USA. ACM.

\bibitem[\protect\citename{Esuli and Sebastiani}2006]{Esuli2006sentiwordnet}
Andrea Esuli and Fabrizio Sebastiani.
\newblock 2006.
\newblock Sentiwordnet: A publicly available lexical resource for opinion
  mining.
\newblock In {\em In Proceedings of the 5th Conference on Language Resources
  and Evaluation (LREC’06}, pages 417--422.

\bibitem[\protect\citename{Farkas \bgroup et al.\egroup
  }2010]{Farkas:2010:CST:1870535.1870536}
Rich\'{a}rd Farkas, Veronika Vincze, Gy\"{o}rgy M\'{o}ra, J\'{a}nos Csirik, and
  Gy\"{o}rgy Szarvas.
\newblock 2010.
\newblock The conll-2010 shared task: Learning to detect hedges and their scope
  in natural language text.
\newblock In {\em Proceedings of the Fourteenth Conference on Computational
  Natural Language Learning --- Shared Task}, CoNLL '10: Shared Task, pages
  1--12, Stroudsburg, PA, USA. Association for Computational Linguistics.

\bibitem[\protect\citename{Feng \bgroup et al.\egroup
  }2013]{conf/acl/FengKKC13}
Song Feng, Jun~Seok Kang, Polina Kuznetsova, and Yejin Choi.
\newblock 2013.
\newblock Connotation lexicon: A dash of sentiment beneath the surface meaning.
\newblock In {\em ACL}, pages 1774--1784. The Association for Computer
  Linguistics.

\bibitem[\protect\citename{Ferschke \bgroup et al.\egroup
  }2012]{Ferschke2012BAR}
Oliver Ferschke, Iryna Gurevych, and Yevgen Chebotar.
\newblock 2012.
\newblock Behind the article: Recognizing dialog acts in wikipedia talk pages.
\newblock In {\em Proceedings of the 13th Conference of the European Chapter of
  the Association for Computational Linguistics}, EACL '12, pages 777--786,
  Stroudsburg, PA, USA. Association for Computational Linguistics.

\bibitem[\protect\citename{Galley \bgroup et al.\egroup }2004]{GalleyEtal}
Michel Galley, Kathleen McKeown, Julia Hirschberg, and Elizabeth Shriberg.
\newblock 2004.
\newblock {Identifying agreement and disagreement in conversational speech: use
  of Bayesian networks to model pragmatic dependencies}.
\newblock In {\em ACL '04: Proceedings of the 42nd Annual Meeting on
  Association for Computational Linguistics}, pages 669+, Morristown, NJ, USA.
  Association for Computational Linguistics.

\bibitem[\protect\citename{Hahn \bgroup et al.\egroup
  }2006]{hahn-ladner-ostendorf:2006:HLT-NAACL06-Short}
Sangyun Hahn, Richard Ladner, and Mari Ostendorf.
\newblock 2006.
\newblock Agreement/disagreement classification: Exploiting unlabeled data
  using contrast classifiers.
\newblock In {\em Proceedings of the Human Language Technology Conference of
  the NAACL, Companion Volume: Short Papers}, pages 53--56, New York City, USA,
  June. Association for Computational Linguistics.

\bibitem[\protect\citename{Hassan \bgroup et al.\egroup
  }2010]{Hassan:2010:WAI:1870658.1870779}
Ahmed Hassan, Vahed Qazvinian, and Dragomir Radev.
\newblock 2010.
\newblock What's with the attitude?: Identifying sentences with attitude in
  online discussions.
\newblock In {\em Proceedings of the 2010 Conference on Empirical Methods in
  Natural Language Processing}, EMNLP '10, pages 1245--1255, Stroudsburg, PA,
  USA. Association for Computational Linguistics.

\bibitem[\protect\citename{Hassan \bgroup et al.\egroup
  }2012]{Hassan:2012:DSO:2390948.2390956}
Ahmed Hassan, Amjad Abu-Jbara, and Dragomir Radev.
\newblock 2012.
\newblock Detecting subgroups in online discussions by modeling positive and
  negative relations among participants.
\newblock In {\em Proceedings of the 2012 Joint Conference on Empirical Methods
  in Natural Language Processing and Computational Natural Language Learning},
  EMNLP-CoNLL '12, pages 59--70, Stroudsburg, PA, USA. Association for
  Computational Linguistics.

\bibitem[\protect\citename{Hirschberg and
  Litman}1993]{Hirschberg:1993:ESD:972487.972490}
Julia Hirschberg and Diane Litman.
\newblock 1993.
\newblock Empirical studies on the disambiguation of cue phrases.
\newblock {\em Comput. Linguist.}, 19(3):501--530, September.

\bibitem[\protect\citename{Joachims}1999]{Joachims:1999:MLS:299094.299104}
Thorsten Joachims.
\newblock 1999.
\newblock Advances in kernel methods.
\newblock chapter Making Large-scale Support Vector Machine Learning Practical,
  pages 169--184. MIT Press, Cambridge, MA, USA.

\bibitem[\protect\citename{Joshi and
  Penstein-Ros{\'e}}2009]{Joshi:2009:GDF:1667583.1667680}
Mahesh Joshi and Carolyn Penstein-Ros{\'e}.
\newblock 2009.
\newblock Generalizing dependency features for opinion mining.
\newblock In {\em Proceedings of the ACL-IJCNLP 2009 Conference Short Papers},
  ACLShort '09, pages 313--316, Stroudsburg, PA, USA. Association for
  Computational Linguistics.

\bibitem[\protect\citename{Lafferty \bgroup et al.\egroup
  }2001]{Lafferty:2001:CRF}
John~D. Lafferty, Andrew McCallum, and Fernando C.~N. Pereira.
\newblock 2001.
\newblock Conditional random fields: Probabilistic models for segmenting and
  labeling sequence data.
\newblock In {\em Proceedings of the Eighteenth International Conference on
  Machine Learning}, ICML '01, pages 282--289, San Francisco, CA, USA. Morgan
  Kaufmann Publishers Inc.

\bibitem[\protect\citename{Mao and Lebanon}2007]{Mao+Lebanon:07a}
Yi~Mao and Guy Lebanon.
\newblock 2007.
\newblock Isotonic conditional random fields and local sentiment flow.
\newblock In {\em Advances in Neural Information Processing Systems}.

\bibitem[\protect\citename{McCallum}2002]{McCallumMALLET}
Andrew~Kachites McCallum.
\newblock 2002.
\newblock Mallet: A machine learning for language toolkit.
\newblock http://mallet.cs.umass.edu.

\bibitem[\protect\citename{Misra and Walker}2013]{misra-walker:2013:SIGDIAL}
Amita Misra and Marilyn Walker.
\newblock 2013.
\newblock Topic independent identification of agreement and disagreement in
  social media dialogue.
\newblock In {\em Proceedings of the SIGDIAL 2013 Conference}, pages 41--50,
  Metz, France, August. Association for Computational Linguistics.

\bibitem[\protect\citename{Qiu \bgroup et al.\egroup
  }2013]{qiu-yang-jiang:2013:NAACL-HLT}
Minghui Qiu, Liu Yang, and Jing Jiang.
\newblock 2013.
\newblock Mining user relations from online discussions using sentiment
  analysis and probabilistic matrix factorization.
\newblock In {\em Proceedings of the 2013 Conference of the North American
  Chapter of the Association for Computational Linguistics: Human Language
  Technologies}, pages 401--410, Atlanta, Georgia, June. Association for
  Computational Linguistics.

\bibitem[\protect\citename{Rashmi~Prasad and Webber}2008]{penntreediscourse}
Alan Lee Eleni Miltsakaki Livio Robaldo Aravind~Joshi Rashmi~Prasad,
  Nikhil~Dinesh and Bonnie Webber.
\newblock 2008.
\newblock The penn discourse treebank 2.0.
\newblock In Bente Maegaard Joseph Mariani Jan Odijk Stelios Piperidis
  Daniel~Tapias Nicoletta Calzolari (Conference~Chair), Khalid~Choukri, editor,
  {\em Proceedings of the Sixth International Conference on Language Resources
  and Evaluation (LREC'08)}, Marrakech, Morocco, may. European Language
  Resources Association (ELRA).
\newblock http://www.lrec-conf.org/proceedings/lrec2008/.

\bibitem[\protect\citename{Somasundaran and
  Wiebe}2009]{Somasundaran:2009:RSO:1687878.1687912}
Swapna Somasundaran and Janyce Wiebe.
\newblock 2009.
\newblock Recognizing stances in online debates.
\newblock In {\em Proceedings of the Joint Conference of the 47th Annual
  Meeting of the ACL and the 4th International Joint Conference on Natural
  Language Processing of the AFNLP: Volume 1 - Volume 1}, ACL '09, pages
  226--234, Stroudsburg, PA, USA. Association for Computational Linguistics.

\bibitem[\protect\citename{Somasundaran and
  Wiebe}2010]{Somasundaran:2010:RSI:1860631.1860645}
Swapna Somasundaran and Janyce Wiebe.
\newblock 2010.
\newblock Recognizing stances in ideological on-line debates.
\newblock In {\em Proceedings of the NAACL HLT 2010 Workshop on Computational
  Approaches to Analysis and Generation of Emotion in Text}, CAAGET '10, pages
  116--124, Stroudsburg, PA, USA. Association for Computational Linguistics.

\bibitem[\protect\citename{Stone \bgroup et al.\egroup }1966]{stone66}
Philip~J. Stone, Dexter~C. Dunphy, Marshall~S. Smith, and Daniel~M. Ogilvie.
\newblock 1966.
\newblock {\em The General Inquirer: A Computer Approach to Content Analysis}.
\newblock MIT Press, Cambridge, MA.

\bibitem[\protect\citename{Thomas \bgroup et al.\egroup
  }2006]{Thomas:2006:GOV:1610075.1610122}
Matt Thomas, Bo~Pang, and Lillian Lee.
\newblock 2006.
\newblock Get out the vote: Determining support or opposition from
  congressional floor-debate transcripts.
\newblock In {\em Proceedings of the 2006 Conference on Empirical Methods in
  Natural Language Processing}, EMNLP '06, pages 327--335, Stroudsburg, PA,
  USA. Association for Computational Linguistics.

\bibitem[\protect\citename{Velikovich \bgroup et al.\egroup
  }2010]{Velikovich2010VWP}
Leonid Velikovich, Sasha Blair-Goldensohn, Kerry Hannan, and Ryan McDonald.
\newblock 2010.
\newblock The viability of web-derived polarity lexicons.
\newblock In {\em Human Language Technologies: The 2010 Annual Conference of
  the North American Chapter of the Association for Computational Linguistics},
  HLT '10, pages 777--785, Stroudsburg, PA, USA. Association for Computational
  Linguistics.

\bibitem[\protect\citename{Walker \bgroup et al.\egroup }2012a]{WALKER12}
Marilyn Walker, Jean~Fox Tree, Pranav Anand, Rob Abbott, and Joseph King.
\newblock 2012a.
\newblock A corpus for research on deliberation and debate.
\newblock In {\em Proceedings of the Eight International Conference on Language
  Resources and Evaluation (LREC'12)}, Istanbul, Turkey, may. European Language
  Resources Association (ELRA).

\bibitem[\protect\citename{Walker \bgroup et al.\egroup }2012b]{WalkerAAG12}
Marilyn~A. Walker, Pranav Anand, Rob Abbott, and Ricky Grant.
\newblock 2012b.
\newblock Stance classification using dialogic properties of persuasion.
\newblock In {\em HLT-NAACL}, pages 592--596. The Association for Computational
  Linguistics.

\bibitem[\protect\citename{Wang \bgroup et al.\egroup }2011]{Wang:2011:DAD}
Wen Wang, Sibel Yaman, Kristin Precoda, Colleen Richey, and Geoffrey Raymond.
\newblock 2011.
\newblock Detection of agreement and disagreement in broadcast conversations.
\newblock In {\em Proceedings of the 49th Annual Meeting of the Association for
  Computational Linguistics: Human Language Technologies: Short Papers - Volume
  2}, HLT '11, pages 374--378, Stroudsburg, PA, USA. Association for
  Computational Linguistics.

\bibitem[\protect\citename{Wilson \bgroup et al.\egroup }2005]{Wilson:2005:RCP}
Theresa Wilson, Janyce Wiebe, and Paul Hoffmann.
\newblock 2005.
\newblock Recognizing contextual polarity in phrase-level sentiment analysis.
\newblock In {\em Proceedings of the Conference on Human Language Technology
  and Empirical Methods in Natural Language Processing}, HLT '05, pages
  347--354, Stroudsburg, PA, USA. Association for Computational Linguistics.

\bibitem[\protect\citename{Yin \bgroup et al.\egroup
  }2012]{Yin:2012:ULG:2392963.2392978}
Jie Yin, Paul Thomas, Nalin Narang, and Cecile Paris.
\newblock 2012.
\newblock Unifying local and global agreement and disagreement classification
  in online debates.
\newblock In {\em Proceedings of the 3rd Workshop in Computational Approaches
  to Subjectivity and Sentiment Analysis}, WASSA '12, pages 61--69,
  Stroudsburg, PA, USA. Association for Computational Linguistics.

\bibitem[\protect\citename{Zhu and Ghahramani}2002]{citeulike:310457}
Xiaojin Zhu and Zoubin Ghahramani.
\newblock 2002.
\newblock Learning from labeled and unlabeled data with label propagation.
\newblock In {\em Technical Report CMU-CALD-02-107}.

\end{thebibliography}

\end{document}